\documentclass[12pt, a4paper]{article}       
\usepackage{fullpage}

\usepackage{hyperref}
\usepackage{pagecolor}
\usepackage{booktabs}
\usepackage{cite}
\usepackage[font={small}]{caption}
\usepackage{subcaption}

\usepackage{array}
\newcolumntype{L}[1]{>{\raggedright\let\newline\\\arraybackslash\hspace{0pt}}m{#1}}
\newcolumntype{C}[1]{>{\centering\let\newline\\\arraybackslash\hspace{0pt}}m{#1}}
\newcolumntype{R}[1]{>{\raggedleft\let\newline\\\arraybackslash\hspace{0pt}}m{#1}}

\let\oldFootnote\footnote
\newcommand\nextToken\relax
\renewcommand\footnote[1]{%
    \oldFootnote{#1}\futurelet\nextToken\isFootnote}
\newcommand\isFootnote{%
    \ifx\footnote\nextToken\textsuperscript{,}\fi}

\usepackage{graphicx} 
\usepackage{epsfig} 
\usepackage{amsmath} 

\newcommand{\myvector}[1]{\overrightarrow{#1}}

\title{
Geometric features for voxel-based surface recognition
 }

\author{Dmitry Yarotsky%
\footnote{Skolkovo Institute of Science and Technology, Skolkovo Innovation Center, Building 3, Moscow  143026, Russia,
        {\tt\small d.yarotsky~at~skoltech.ru}}
\footnote{Institute for Information Transmission Problems, Bolshoy Karetny per. 19, build.1, Moscow 127051, Russia
        }%
}
\date{}

\begin{document}

\maketitle

\begin{abstract}
We introduce a library of geometric voxel features for CAD surface  recognition/retrieval tasks. Our features include local versions of the intrinsic volumes (the usual 3D volume, surface area, integrated mean and Gaussian curvature) and a few closely related quantities. We also compute Haar wavelet and statistical distribution features by aggregating raw voxel features. We apply our features to object classification on the ESB data set and demonstrate accurate results with a small number of shallow decision trees.

\end{abstract}


\section{Introduction}

Automated classification of CAD shapes, along with retrieval of similar shapes, is an old and well-studied area of research. The common approach to automated classification/retrieval consists in first representing the shape by a \emph{feature vector} using some collection of descriptors, and then, for the classification task, training a classifier using some general classification algorithm applicable to array-like data, or, for the retrieval task, comparing feature vectors and finding nearby vectors in the feature space. Here, we use the notion of feature vector in a broad sense, as any standardized, fixed-size representation of the object. Replacing the original description by a feature vector serves the important purposes of reducing the complexity of the representation and making all the (otherwise possibly quite diverse) objects directly comparable with each other.  There is a large variety of very different strategies to form the feature vector; see 3D retrieval surveys \cite{bustos2005feature, tangelder2008survey, bimbo2006content, zhang2007survey, iyer2005three, cardone2003survey}. 

A simple class of features is global geometric properties such as the (properly rescaled) total area, volume, low degree moments, or Euler characteristic \cite{mecke2000additivity, zhang2001efficient, corney2002coarse}. These global features are appealing thanks to their direct geometric meaning, but are relatively crude to distinguish complex shapes.  

Another popular class of features comes from statistical  description of local properties. In this approach, feature vectors can be thought of as histograms of quantities characterizing the local surface geometry (e.g., orientation, point-to-point distance, curvature, etc.) on a certain length scale  \cite{osada2002shape,horn1984extended,ankerst19993d}. 

Recently, it has become very popular to approach the classification/retrieval tasks using deep learning and surface voxelization \cite{wu20153d,graham2014spatially,brock2016generative,DBLP:journals/corr/NotchenkoKB16}. In this case, the feature vector is the assignment of Boolean values (``empty/occupied'') to the set of all voxels. In contrast to some other feature generation strategies, voxelization preserves information on spatial location of surface elements, which allows one to combine this approach with localization-aware learning methods such as convolutional neural networks.  

The purpose of the present paper is to describe a unified framework combining voxel representation of CAD models with standard geometric properties such as area, curvature, orientation, etc. Specifically, instead of just marking a voxel as ``empty'' or ``occupied'', we equip it with a detailed description of the local surface geometry. While this general idea is certainly not new (see, e.g., \cite{vranic20013d, kriegel2003using, kriegel2003effective, godil2011salient}), some elements of our approach distinguish it from earlier research.

\begin{itemize}
\item In each voxel, we aim to obtain a possibly rich yet consistent and compact characterization of the local geometry. In particular, our features include voxel-restricted versions of all four 3D morphological Minkowski functionals \cite{mecke2000additivity,santalo2004integral}. This also allows us to directly compare efficiency of different feature types. 
\item We aim to consider geometric features on different length scales in a unified fashion, mostly through the additivity property (the value of the feature on the larger scale is the sum of the corresponding values on the smaller scale). When evaluated on the largest scale, some of our features become familiar global quantities: total area, volume, the complete integral of mean curvature, and Euler characteristic. This allows us to directly compare efficiency of different scales for feature representation.  
\item We explicitly decouple feature generation in voxels from subsequent post-processing and learning phases. This allows us to directly evaluate effects of different feature aggregation strategies, e.g., the Haar wavelet transform and histograms of feature values.
\item The natural trend in voxelization-based 3D shape classification/retrieval is the increase in resolution, thanks to the constantly growing number and diversity of 3D models. As CAD models are essentially 2D surfaces, only a small share of voxels is occupied at high resolutions. Our implementation of voxel features is sparse-vector-based, which allows us to consider resolutions $N\times N\times N$ with $N>100$. 
\end{itemize}

The code for this work has been open-sourced as the library {\bf voxelfeatures}.\footnote{\url{https://github.com/yarotsky/voxelfeatures/}} 
The library has a C++ feature-generation core and Python bindings, and can be installed in a conventional way as an importable Python module.

\begin{figure*}[thb] 
\begin{center}
\includegraphics[scale=0.47, clip, trim=5mm 10mm 10mm 20mm]{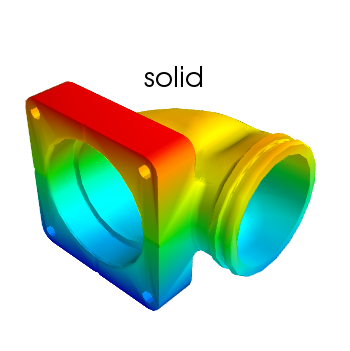}
\includegraphics[scale=0.47]{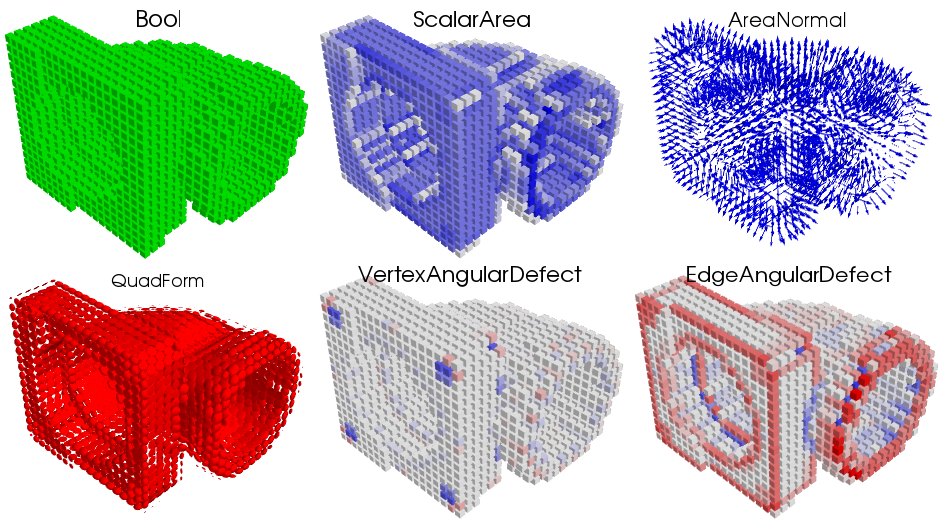}
\caption{The original surface and six voxelizations with different local features.}\label{fig:feature_examples}
\end{center}
\end{figure*}

\section{Voxel features}\label{sec:raw}

The main focus of our work is a library of local features computed in each voxel of the rasterized representation. Given a surface $\Sigma$ and a cubic voxel $V$, a local feature is some quantity evaluated for the intersection $\Sigma\cap V$ and characterizing some geometric aspects of  this piece of the surface: orientation, roughness, curvature, etc. 

Our selection of features is primarily influenced by the integral geometry point of view. Our first choice is the voxel-restricted family of intrinsic volumes (also known by many other names: valuations,  Minkowski functionals, etc.). In 3D, there are four intrinsic volumes: the usual 3D volume, the surface area, and the integrated mean and Gaussian curvatures of the surface \cite{santalo2004integral}. The intrinsic volumes are known to generally play a major role in practical pattern recognition problems \cite{vogel2010quantification, mecke2000additivity}.

One important property of the instrinsic volumes is their Hausdorff-continuity on convex sets. Smooth surfaces $\Sigma$ are commonly represented in practice by triangle meshes $\Sigma'$ approximating them with some accuracy. We are interested in those geometric features that are not too sensitive to such an approximation. In particular, a reasonable feature should not depend on the way in which a polygonal face is divided into triangles. One way to formulate a general requirement is to say that the feature should be continuous w.r.t. the Hausdorff distance between surfaces, $$\operatorname{dist}(\Sigma, \Sigma')=\max(\max_{\mathbf x\in \Sigma}\min_{\mathbf y\in\Sigma'} |\mathbf{x}-\mathbf{y}|, \max_{\mathbf x\in \Sigma'}\min_{\mathbf y\in\Sigma} |\mathbf{x}-\mathbf{y}|).$$
The intrinsic volumes are Hausdorff-continuous on the set of surfaces bounding convex bodies (though not on arbitrary surfaces, in general). 

Another important property of the intrinsic volumes is their Euclidean invariance (invariance with respect to rigid motions).

Finally, the intrinsic volumes are additive, in the sense that $F(A)+F(B)=F(A\cap B)+ F(A\cup B),$ where $F(A)$ denotes an intrinsic volume of a body $A$.

The central general result in integral geometry, Hadwiger's theorem, states that 
any additive, Euclidean invariant functional on $d$-dimensional bodies that is Hausdorff-continuous on the set of convex bodies is a linear combination of the $d+1$ intrinsic volumes (in particular, the four aforementioned volumes in the 3D case) \cite{chen2004simplified,hadwiger2013vorlesungen,klain1995short}.

We slightly adapt the concept of intrinsic volumes for our purpose of defining voxel features, also taking into account the fact that computationally we work with surfaces rather than bodies. Namely, all the intrinsic volumes except for the 3D volume are  surface integrals of elementary symmetric polynomials of the local curvatures; we define the respective voxel features (see the features SA, VAD, EAD below) by restricting the surface integral to the voxel. Since the 3D volume can also be written as a surface integral, we define the respective feature (see VE below) in the same fashion, by restricting this surface integral to the voxel. 

Clearly, due to the Euclidean symmetry breaking and boundary effects associated with voxelization, the resulting voxel features do not retain the Euclidean invariance and continuity of the intrinsic volumes; however, we expect that our choice of features should in some sense minimize these negative effects. The additivity property is retained in the sense that if a voxel is represented as a union of disjoint voxels on a lower length scale, $V=\cup_k V_k,$ then $F(V)=\sum_k F(V_k)$. In particular, given a conventional single  length scale voxelization, the total sum of the feature values in all voxels gives the full intrinsic volume. 

The intrinsic volumes are orientation-independent, which is useful for rotation-invariant object classes, but in some cases information about orientation is important (e.g., to distinguish object's top from bottom). To this end, we also consider some features containing this information and defined, as before, by integrating over the surface restricted to the voxel (see AN, QF below).

As the discussion above shows, we are mostly interested in the case when the surface $\Sigma$ is a boundary of a solid body. In this case the triangle mesh $\Sigma'$ is supposed to be consistent, in the sense that each edge is incident to exactly two faces, and the orientations of these two faces are compatible. Such consistent meshes are sometimes referred to as ``watertight'', while unconstrained collections of triangles are referred to as ``polygon soups''. Some of our voxel features, but not all, require consistent meshes.

Computation of our features is based on first finding the polygonal decomposition of $\Sigma'\cap V$ for each voxel $V$, which is done recursively by octree traversal; only non-empty voxels are analyzed so that the full feature vector is a sparse vector.

Below we list our features, primarily defining them in general integral-geometric terms, and detailing how they are computed from the polygonal mesh only where it is not obvious (otherwise, integrals are replaced by respective sums, etc.). Some properties of the features are summarized in Table \ref{tab:features}  and examples are shown in Fig. \ref{fig:feature_examples}.

\begin{table}[tpb]
\begin{center}
\begin{tabular}{l c C{2.7cm} c}
\toprule
Feature & Dim & Assumes mesh consistency & Additive\\
\midrule
Bool & 1 & No & No\\
SA & 1 & No & Yes \\
AN & 3 & Yes & Yes\\
QF & 6 & No & Yes\\
EV & 3 & No & No\\
VAD & 1 & Yes & Yes\\
EAD & 1 & Yes & Yes\\
VE & 1 & Yes & Yes\\
\bottomrule
\end{tabular}
\caption{Properties of voxel features}
\label{tab:features}
\end{center}
\end{table}

\paragraph{Bool} 

$$\operatorname{Bool}(\Sigma,V) = \begin{cases}1, & \Sigma\cap V=\emptyset\\0, & \text{otherwise}\end{cases}$$
This is the trivial feature most often used in voxelization. In contrast to other features, Bool can in principle be computed without finding the detailed polygonal decomposition of $\Sigma'\cap V$.   

\paragraph{SurfaceArea (SA)} 

$$\operatorname{SA}(\Sigma,V) = \int_{\Sigma\cap V}dS$$
Obviously, the total value of this additive feature (the value at the trivial $1\times 1\times 1$ voxelization) is the total area of the surface.

\paragraph{AreaNormal (AN)} 

$$\operatorname{AN}(\Sigma,V) = \int_{\Sigma\cap V}\myvector{\mathbf{n}}dS,$$
where $\myvector{\mathbf{n}}$ is the local normal to the surface. This additive feature requires consistent meshes with a fixed surface orientation (with all normals either outward or inward). For a consistent mesh the total value of AN is $\mathbf{0}$.

\paragraph{QuadForm (QF)} 

$$\operatorname{QF}(\Sigma,V) = \int_{\Sigma\cap V}\myvector{\mathbf{n}} \myvector{\mathbf{n}}^tdS,$$
where $\myvector{\mathbf{n}} \myvector{\mathbf{n}}^t$ is the outer product, so that QF is a symmetric positive-definite $3\times 3$ matrix (with 6 independent parameters). In contrast to AN, this feature is not affected by reversing the normal direction and so does not require mesh consistency. The trace of QF equals SA.

\paragraph{EigenValues (EV)} 

This feature represents the three sorted eigenvalues of the positive-definite matrix QF defined above. EV contains the rotation-invariant information about QF, which makes it more suitable than QF for orientation independent tasks. 

\paragraph{VertexAngularDefect (VAD)}
This feature represents the integral of the Gaussian curvature:
$$\operatorname{VAD}(\Sigma,V)=\int_{\Sigma\cap V}k_1k_2dS,$$
where $k_1,k_2$ are the two principal curvatures at surface points.

This feature requires a consistent mesh. Given such a mesh, VAD can be expressed as the total angular defect of all vertices found in the voxel $V$. The angular defect of a vertex is defined as $2\pi-\sum \alpha_n$, where $\alpha_n$ are the incident angles of the polygonal surface. This angular defect is the polyhedral analog of the Gaussian curvature: if a smooth surface is approximated by polyhedra, then the total angular defect in a given domain will converge to the integral of the Gaussian curvature in this domain (conversely, if a polyhedral surface is approximated by smooth surfaces, then the Gaussian curvature will concentrate at the vertices of the polyhedral surface). By Descartes' theorem (or Gauss-Bonnet theorem in the smooth case), the total angular defect is a topological invariant and equals $2\pi$ times the Euler characteristic of the surface ($2-2\cdot\text{``number of handles''}$, assuming that the surface is connected). 

\paragraph{EdgeAngularDefect (EAD)}
This feature represents the integral of the mean curvature (with the factor 2):
$$\operatorname{EAD}(\Sigma,V)=\int_{\Sigma\cap V}(k_1+k_2)dS.$$
The version of this formula for a polygonal surface reads 
$$\operatorname{EAD}(\Sigma',V)=\sum_{\text{edges } e}(\pi-\beta_e)l_{e\cap V},$$
where $\beta_e$ is the dihedral angle at the edge $e$, and $l_{e\cap V}$ is the length of the intersection of the edge $e$ with the voxel $V$. 

\paragraph{VolumeElement (VE)}
$$\operatorname{VE}(\Sigma,V)=\frac{1}{3}\int_{\Sigma\cap V}\myvector{\mathbf{r}}^t \myvector{\mathbf{n}}dS,$$
where $\myvector{\mathbf{r}}$ is the radius vector and $\myvector{\mathbf{r}}^t \myvector{\mathbf{n}}$ the inner product of this vector with the normal. Assuming that the normal is outward, the total value of VE is the volume of the body bounded by the surface $\Sigma$. In contrast to other features, VE is not translation-invariant (as $\myvector{\mathbf{r}}$ depends on the position of the origin).

\section{Feature aggregation}\label{sec:agg}
At high resolutions, the local features computed in voxels as described above tend to produce an excessively detailed description of the surface. A classifier trained on such a description will likely overtrain substantially. One way to counteract this is to use a classification algorithm with overtraining prevention mechanisms -- e.g., a convnet, as it utilizes weight sharing. Another way is to directly transform the initial, ``raw'' voxel features into some more appropriate ``aggregated'' features. The available grid-like structure of the raw voxel representation is essential for some aggregation strategies.  

One example of such aggregation is a wavelet transformation of the voxel data. We consider the simplest case of Haar wavelets \cite{haar1910theorie}. Suppose that a scalar voxel feature is evaluated at a resolution $N=2^n$, so that the total feature space can be  written as $W\otimes W\otimes W,$ where $W$ is the $2^n$-dimensional space of feature values along one coordinate. The Haar transform can then be written as $H\otimes H\otimes H$, where $H$ is the orthogonal $2^n\times 2^n$ matrix with the entries
$$H_{k, s}=\begin{cases}
2^{-n/2}, & k=1,\\
2^{-(n-m)/2}, & 
2^{m}<k\le 2^{m+1},
(k-2^m-1)2^{n-m}<s
\le (k-2^m-1/2)2^{n-m},
\\
-2^{-(n-m)/2}, & 
2^{m}<k\le 2^{m+1},
(k-2^m-1/2)2^{n-m}<s
\le (k-2^m)2^{n-m},
\\
0, & \text{otherwise},
\end{cases}$$
with $m=0,\ldots, n-1$ characterizing the length scale.
As the Haar transform is linear and orthogonal, the transformed features contain the same information as the original raw features, but this information is now redistributed over several different length scales according to the value of $m$. This explicit exposure of the length scale hierarchy can help reduce overtraining and is an alternative to generating raw voxel features at several resolutions.  

Another example of feature aggregation is to compute a histogram or some other statistical measure of feature values over the non-empty voxels. In the experiments below we consider a few feature percentiles. In particular, the 0'th percentile corresponds to the minimal value, the 100'th to the maximum value, and the 50'th to the median value.

In the case of vector-valued features we perform feature aggregation separately for each component.

\section{Experiments}

\begin{figure*}[thb] 
\begin{center}
\includegraphics[width=\textwidth]{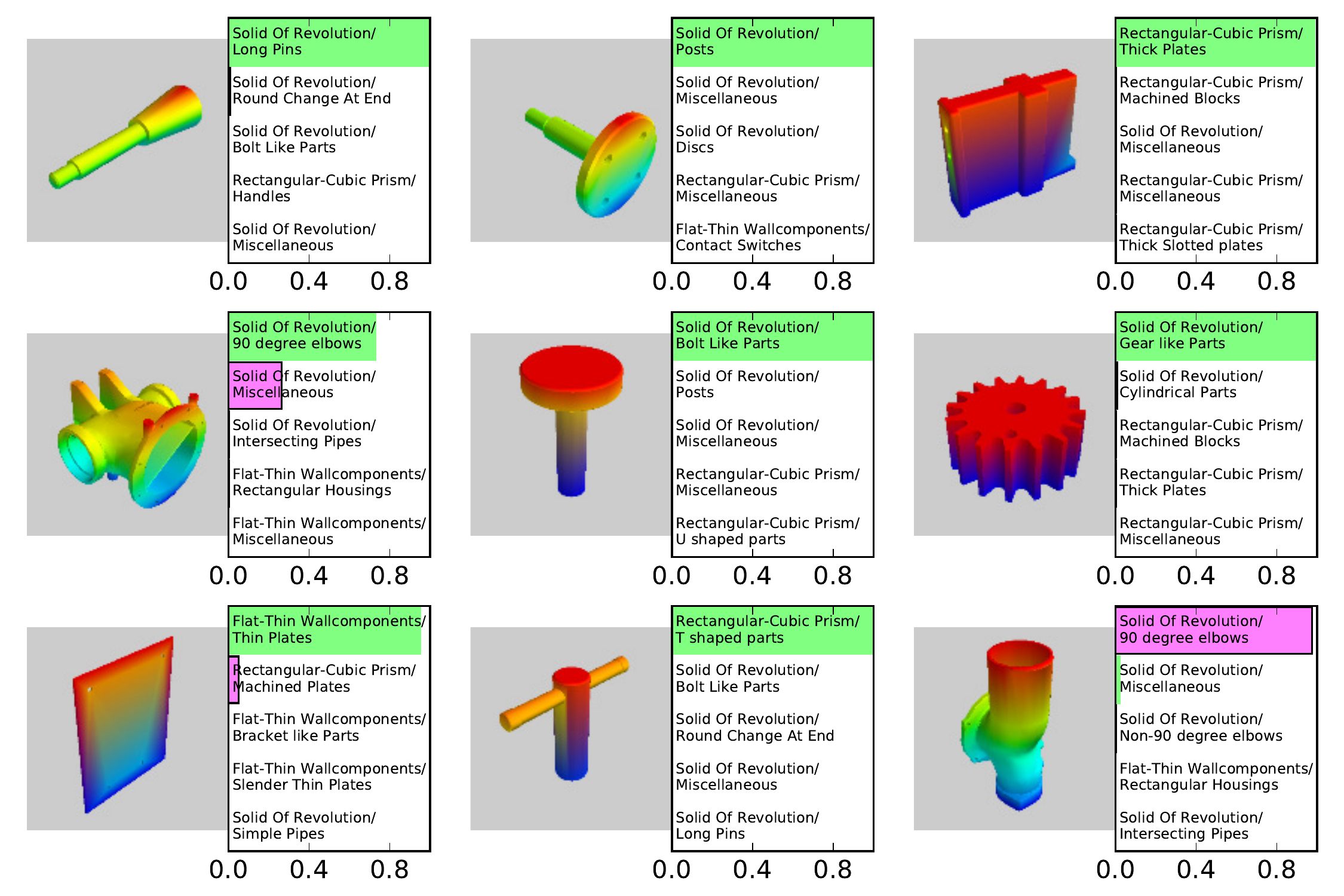}
\caption{Predictions of a classifier with test error 0.15 on several random test objects. For each object, top 5 classes in terms of the predicted probability are shown. The highest probability is correctly assigned to the true class for all objects except for the pipe in the lower right corner (in this case the true class has the second-highest probability). }\label{fig:pred_examples}
\end{center}
\end{figure*}

We have performed a series of classification experiments with the ESB  (Engineering Shape Benchmark, \cite{jayanti2006developing}) collection of CAD models. This collection consists of 866 mechanical parts divided into 3 ``superclasses'' (Flat-thin wall components, Rectangular cubic prisms, and Solids of revolution). Each superclass is further divided into finer classes (like Clips, Handles, Bolt like parts, etc.); there are 45 fine classes in total (hereafter referred to simply as classes). In particular, each superclass contains a Miscellaneous class for objects that cannot be easily assigned to other classes of this superclass. See Fig. \ref{fig:pred_examples} for some examples of ESB objects.

We have divided the collection into a training set of 675 objects and a test set of 191 objects, to be used in all the experiments. A classifier's prediction for a given object is a 45-dimensional vector of class probabilities. We measure the total prediction error simply as the fraction of test objects where the true class is different from the one with the highest predicted probability.

All the ESB models have consistent, watertight meshes, so all the features from Table \ref{tab:features} are applicable.

As the ESB collection is relatively small, classifiers tend to overfit on it substantially. A common strategy to somewhat reduce overfitting is to symmetrize the training set and the classifier. Specifically, if the class of a surface $\Sigma$ is supposed to be invariant under the action of some group of transformations $G$ (for a $g\in G$ we denote the action of $g$ on $\Sigma$ by $g\Sigma$), we can, first, augment the original training set $T$ with the transformed surfaces $\cup_{g\in G}gT$ and, second, symmetrize the predictions of the classifier $C$: $$C_{\mathrm{sym}}(\Sigma)=\frac{1}{|G|}\sum_{g\in G}C(g\Sigma).$$
If $G$ is infinite, or finite but large, one takes a sample of $G$  rather than the full group.

Classification of the ESB models is clearly $O(3)$-invariant, and the models do not, in general, have a preferred orientation. One  way to exploit this invariance is to rotate the surface to its principal axes and then symmetrize over the $2^3\cdot 3!=48$ reflections and permutations of these axes. We have found it, however, to be somewhat more efficient to directly sample the group $O(3)$. The results described below have been obtained with the data and classifiers symmetrized using 20 random rotations followed by centering and rescaling the surfaces to fit the unit cube.

Our classifiers are ensembles of gradient boosted trees trained using the XGBoost library \cite{chen2016xgboost}. While it is common to train voxelization-based classifiers with neural networks \cite{graham2014spatially,wu20153d}, XGBoost has allowed us to easily mix raw and aggregated features as well as features on different length scales, and examine their relative effect and the complexity of the resulting classifiers. 

The experiments were performed on a computer with Intel Core i5-3470 CPU and 16 GB RAM.  

\begin{figure}[tpb] 
\begin{center}
\begin{subfigure}[b]{0.45\textwidth}
\includegraphics[scale=0.6,clip,trim=3mm 3mm 3mm 3mm]{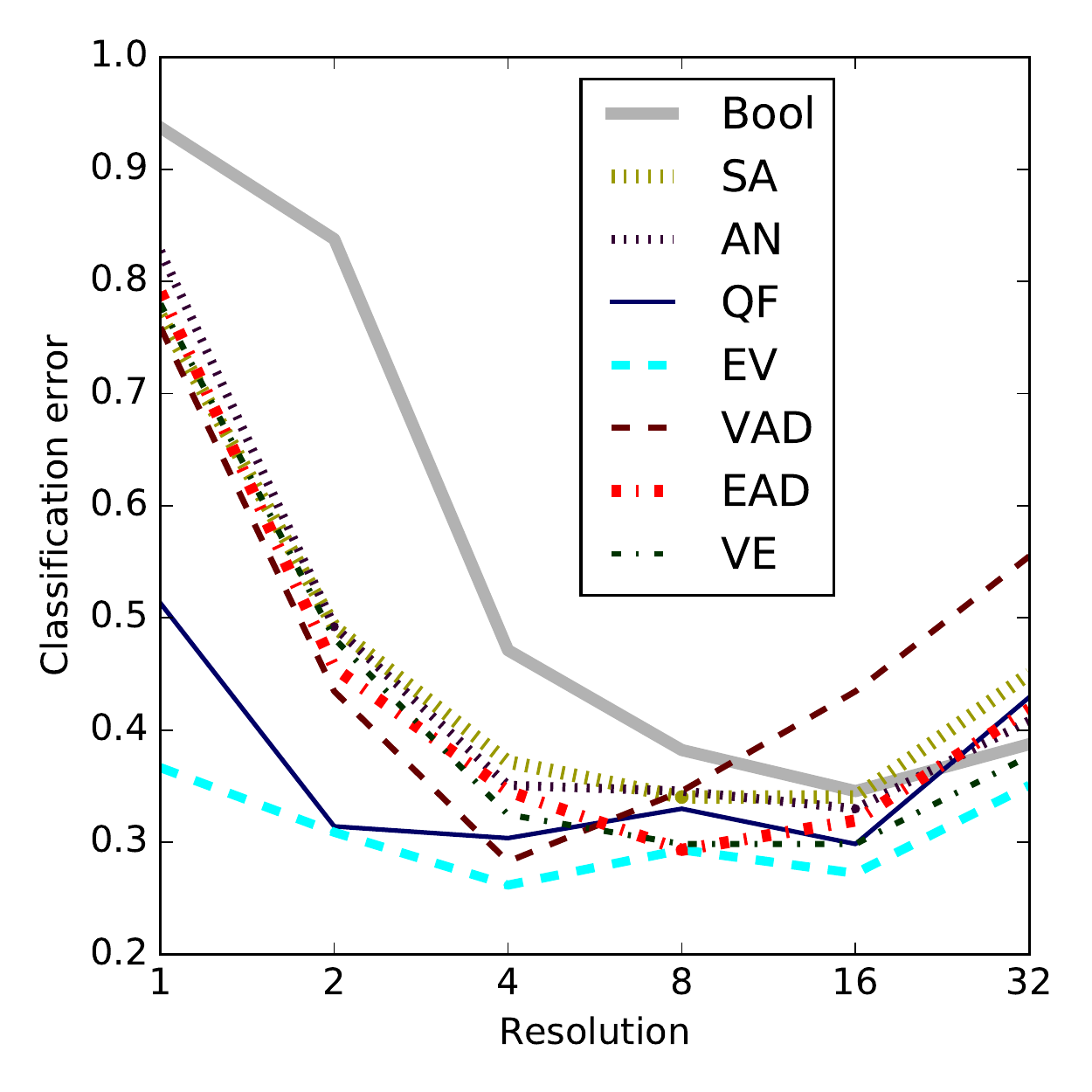}
\caption{}\label{fig:resol_vs_error}
\end{subfigure}\quad
\quad
\begin{subfigure}[b]{0.45\textwidth}
\includegraphics[scale=0.6,clip,trim=3mm 3mm 3mm 3mm]{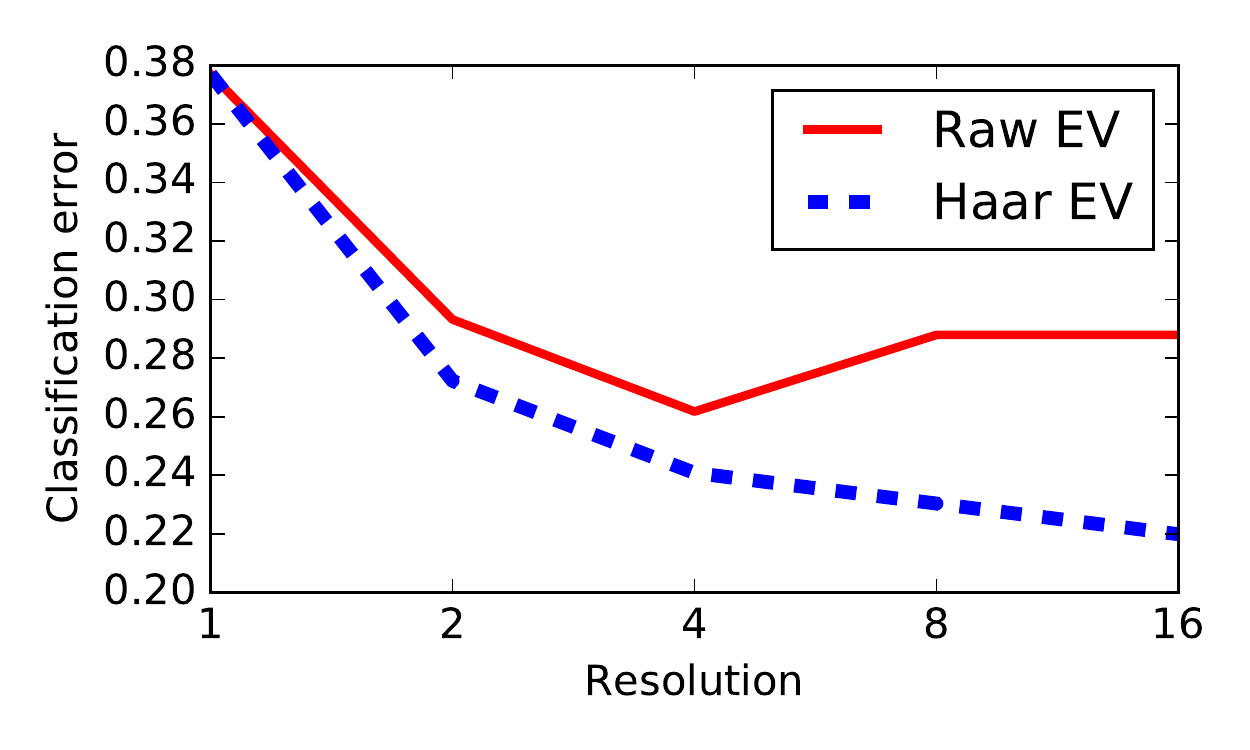}
\caption{}\label{fig:raw_vs_haar}
\end{subfigure}
\caption{(a) Relative efficiency of different features and resolutions for classification on ESB. The classifier is trained by XGBoost, as an ensemble of depth-6 decision trees. The classification features are the raw voxel features (e.g., $3\cdot 4^3=192$ features in the case of EV at resolution 4). (b) Relative efficiency of the raw and Haar-aggregated EV features at different resolutions.}
\end{center}
\end{figure}

In the first series of experiments we compare the efficiency of different individual raw features at different resolutions, see Fig. \ref{fig:resol_vs_error}. The tree depth for these experiments was fixed at the default XGBoost value 6, which was found through preliminary testing to produce good results. We observe that the optimal resolutions are those between 4 and 16: at higher values the voxelized description is too detailed to be efficiently processed by XGBoost.     
Some features, namely QF and EV, turn out to be quite discriminative even at the trivial resolution 1. This is not surprising, since many classes have typical characteristics reflected in the global 3-dimensional feature EV, e.g. pipes and hollow bodies have a relatively large area (which is the sum of the EV values), some classes have a cylindrical symmetry (then two of the EV values are equal), some classes are characterized by an elongated shape (then two of the EV values are much larger than the third) or flatness (one EV value is much larger then the other two). QF is essentially EV with the added orientation data; at the trivial resolution 1 the orientation data is a noise that can only degrade the recognition accuracy, so QF's performance is worse than that of EV. At resolutions higher than 1 the classification accuracy is even higher, though the EV values are then not as easily interpretable. The trivial Bool feature catches up with more complex features at high resolutions, but on the whole is outperformed by other features.

In the second series of experiments we compare raw voxel EV features with the respective Haar wavelet features at different resolutions, see Fig. \ref{fig:raw_vs_haar}. We see that, as discussed in Section \ref{sec:agg}, despite containing the same information, the Haar representation helps prevent overfitting at large resolutions.  

Finally, we construct a classifier using multiple local raw and statistical (percentile) features evaluated at multiple length scales. Specifically, we include in the set of features all raw non-Bool features from Section \ref{sec:raw}  at resolutions 1, 2 and 4, and all (non-Bool) statistical features at the resolutions $\{2^n\}_{n=1}^7$. Note that we can consider statistical voxel features at high resolutions thanks to our sparse representation of voxel features. Our statistical features include the percentiles 0, 25, 50, 75 and 100. It was determined through preliminary experiments that with these features the optimal results are achieved with very shallow, depth-2 trees.

\begin{figure}[tpb] 
\begin{center}
\begin{subfigure}[b]{0.45\textwidth}
\includegraphics[scale=0.6,clip,trim=3mm 3mm 3mm 3mm]{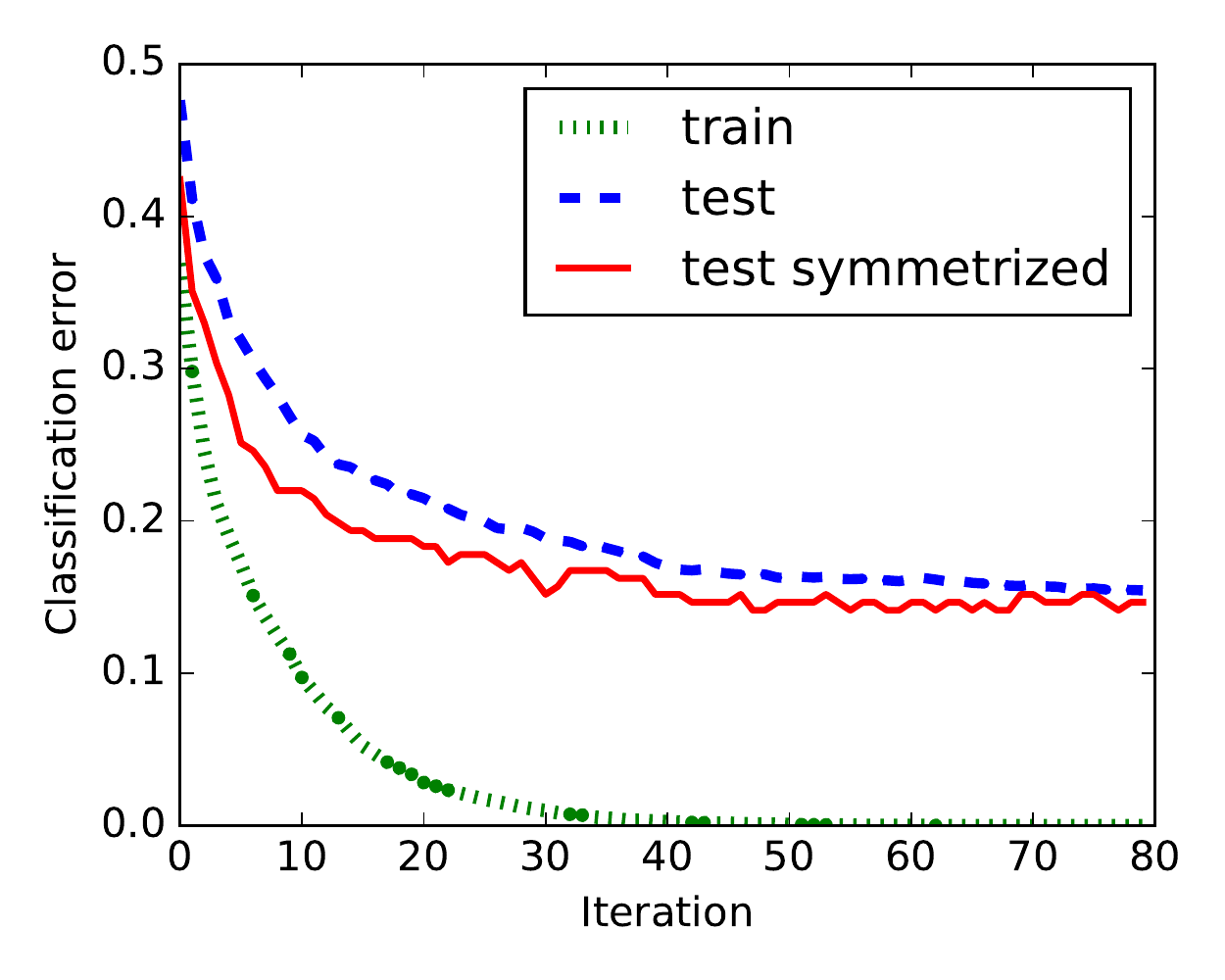}
\caption{}\label{fig:history}
\end{subfigure}\quad
\quad
\begin{subfigure}[b]{0.45\textwidth}
\includegraphics[scale=0.6,clip,trim=3mm 3mm 3mm 3mm]{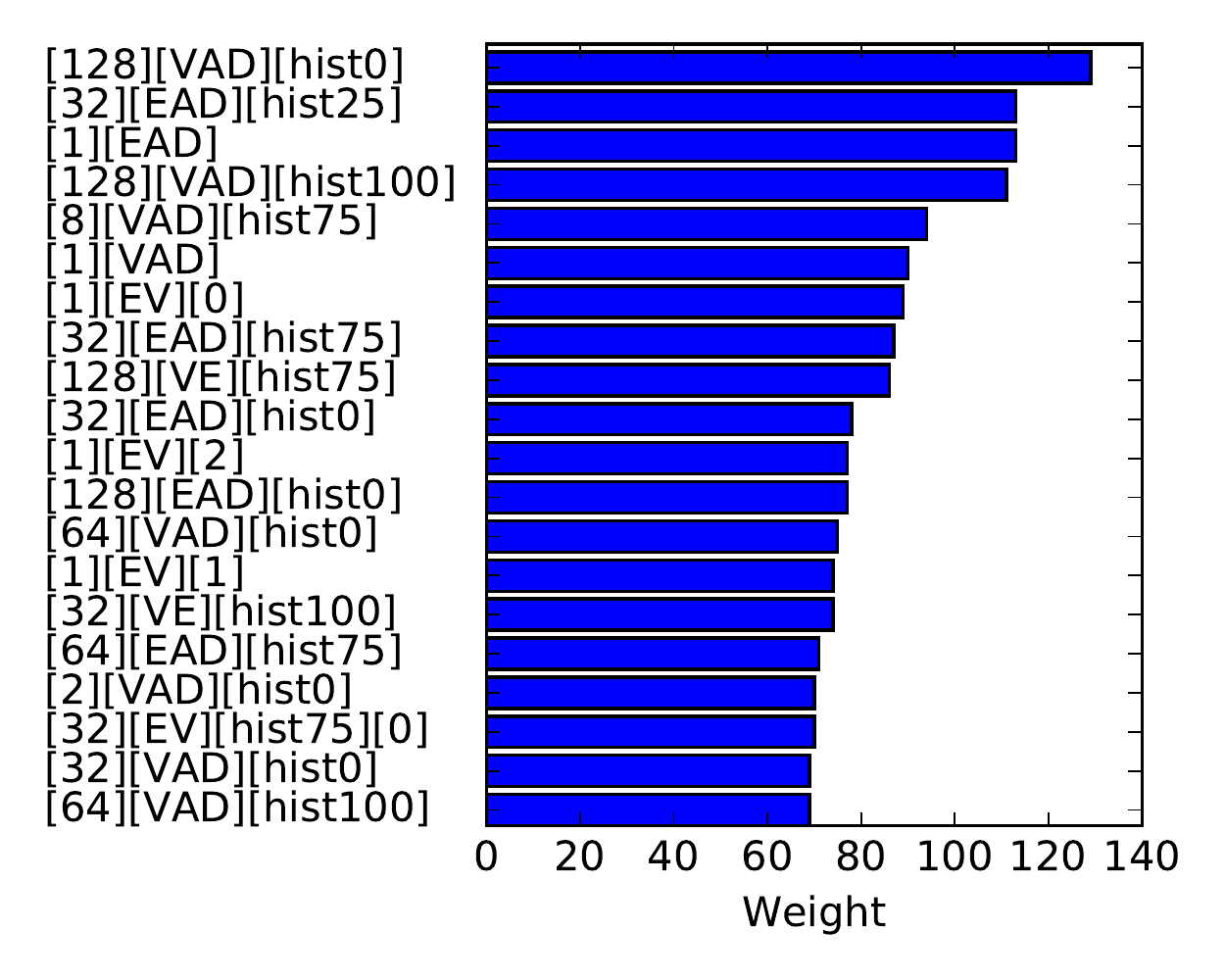}
\caption{}\label{fig:hist}
\end{subfigure}
\caption{(a) The error history for an XGBoost classifier trained using depth-2 decision trees and both raw and statistical features. At each iteration a new tree is added per class. The test errors are shown separately for the symmetrized and non-symmetrized predictions. (b) The top 20 raw and statistical features, in terms of the number of occurrences in the decision trees. The features are shown in the format [Resolution][Feature type][Percentile][Component], where Percentile is indicated with the prefix \texttt{hist} and is only present for statistical features, while Component is only present for non-scalar features (AN, QF, EV).}
\end{center}
\end{figure}

In Fig. \ref{fig:history} we show the error convergence history of the trained classifier. The test error is shown separately for the $O(3)$-symmetrized and non-symmetrized classifiers. We observe, in particular, that with the non-symmetrized classifier it takes about 10 (respectively, 30) depth-2 trees per class to reach the classification errors 0.3 (0.2), while with the symmetrized one it takes 5 (15) trees. At the end of training the test error is approximately 0.15.

In Fig. \ref{fig:hist} we show the top 20 features (both raw and aggregated) in terms of the number of their occurrences in the trees of the final classifier. We observe that the top features include raw features at the trivial resolution 1 as well as statistical features at large resolutions. While the earlier experiment with individual features in Fig. \ref{fig:resol_vs_error} suggests that raw features can be useful at resolutions higher than 1 with deeper trees, in the present experiment they are clearly outperformed by statistical features. We also observe that the classifier especially favors the VAD-, EAD-, and EV-based features and does not favor the SA-, QF- and AN-based ones. Interestingly, neither the total area (\texttt{[1][SA]}) nor the total volume (\texttt{[1][VE]}) is among the top-20 features, though some statistical VE features are. 

In Fig. \ref{fig:pred_examples} we show typical predictions of the classifier. We remark that, not surprisingly, confusions involving the Miscellaneous classes seem to be especially frequent. 

\section{Conclusion}

We have introduced and examined several voxel features inspired by integral geometry, as  well as some more complex (Haar and statistical) features derived from them. We have tested our features on the task of classifying ESB CAD models with gradient boosted trees and observed a reasonably good performance. Our framework has allowed us to compare efficiency of different features and different resolutions at which they are evaluated. In particular, we have found that, within this framework, the best ESB classification results were achieved with very shallow, depth-2 trees that favored high resolution statistical features as well as some global features such as the total integrals of the Gaussian and mean curvature.

Generation of complex features for high resolution voxelization is a relatively complex and time-consuming task, so, speaking about classification, there is some trade-off between using complex features with simple predictive models (as in our case) and using simple features with complex models (as with deep convnets applied to Boolean voxelization). The former alternative may be preferable, for example, if a possibly explicit description or in-depth analysis of the predictive model is desired.

As we have already mentioned, though our voxel features are primarily inspired by the global intrinsic volumes, they, for obvious reasons, do not share the defining properties of the latter, in particular the Euclidean invariance. It would be interesting to develop some general integral-geometric theoretical framework for features compatible with voxelization.


\section*{Acknowledgment}
The author thanks Ermek Kapushev, Alexandr Notchenko and Evgeny Burnaev for many interesting discussions.

\bibliographystyle{plain}
\bibliography{main}

\begin{thebibliography}{10}

\bibitem{ankerst19993d}
Mihael Ankerst, Gabi Kastenm{\"u}ller, Hans-Peter Kriegel, and Thomas Seidl.
\newblock 3d shape histograms for similarity search and classification in
  spatial databases.
\newblock In {\em International Symposium on Spatial Databases}, pages
  207--226. Springer, 1999.

\bibitem{bimbo2006content}
Alberto~Del Bimbo and Pietro Pala.
\newblock {Content-based retrieval of 3D models}.
\newblock {\em ACM Transactions on Multimedia Computing, Communications, and
  Applications (TOMM)}, 2(1):20--43, 2006.

\bibitem{brock2016generative}
Andrew Brock, Theodore Lim, JM~Ritchie, and Nick Weston.
\newblock Generative and discriminative voxel modeling with convolutional
  neural networks.
\newblock {\em arXiv preprint arXiv:1608.04236}, 2016.

\bibitem{bustos2005feature}
Benjamin Bustos, Daniel~A Keim, Dietmar Saupe, Tobias Schreck, and Dejan~V
  Vrani{\'c}.
\newblock Feature-based similarity search in 3d object databases.
\newblock {\em ACM Computing Surveys (CSUR)}, 37(4):345--387, 2005.

\bibitem{cardone2003survey}
Antonio Cardone, Satyandra~K Gupta, and Mukul Karnik.
\newblock A survey of shape similarity assessment algorithms for product design
  and manufacturing applications.
\newblock {\em Journal of Computing and Information Science in Engineering},
  3(2):109--118, 2003.

\bibitem{chen2004simplified}
Beifang Chen.
\newblock {A simplified elementary proof of Hadwiger's volume theorem}.
\newblock {\em Geometriae Dedicata}, 105(1):107--120, 2004.

\bibitem{chen2016xgboost}
Tianqi Chen and Carlos Guestrin.
\newblock {XGBoost: A scalable tree boosting system}.
\newblock {\em arXiv preprint arXiv:1603.02754}, 2016.

\bibitem{corney2002coarse}
Jonathan Corney, Heather Rea, Doug Clark, John Pritchard, Michael Breaks, and
  Roddy MacLeod.
\newblock Coarse filters for shape matching.
\newblock {\em IEEE Computer Graphics and Applications}, 22(3):65--74, 2002.

\bibitem{godil2011salient}
Afzal Godil and Asim~Imdad Wagan.
\newblock Salient local 3d features for 3d shape retrieval.
\newblock In {\em IS\&T/SPIE Electronic Imaging}, pages 78640S--78640S.
  International Society for Optics and Photonics, 2011.

\bibitem{graham2014spatially}
Benjamin Graham.
\newblock Spatially-sparse convolutional neural networks.
\newblock {\em arXiv preprint arXiv:1409.6070}, 2014.

\bibitem{haar1910theorie}
Alfred Haar.
\newblock {Zur Theorie der orthogonalen Funktionensysteme}.
\newblock {\em Mathematische Annalen}, 69(3):331--371, 1910.

\bibitem{hadwiger2013vorlesungen}
Hugo Hadwiger.
\newblock {\em {Vorlesungen {\"u}ber inhalt, Oberfl{\"a}che und
  isoperimetrie}}, volume~93.
\newblock Springer-Verlag, 2013.

\bibitem{horn1984extended}
Berthold Klaus~Paul Horn.
\newblock {Extended Gaussian images}.
\newblock {\em Proceedings of the IEEE}, 72(12):1671--1686, 1984.

\bibitem{iyer2005three}
Natraj Iyer, Subramaniam Jayanti, Kuiyang Lou, Yagnanarayanan Kalyanaraman, and
  Karthik Ramani.
\newblock Three-dimensional shape searching: state-of-the-art review and future
  trends.
\newblock {\em Computer-Aided Design}, 37(5):509--530, 2005.

\bibitem{jayanti2006developing}
Subramaniam Jayanti, Yagnanarayanan Kalyanaraman, Natraj Iyer, and Karthik
  Ramani.
\newblock Developing an engineering shape benchmark for cad models.
\newblock {\em Computer-Aided Design}, 38(9):939--953, 2006.

\bibitem{klain1995short}
Daniel~A Klain.
\newblock {A short proof of Hadwiger's characterization theorem}.
\newblock {\em Mathematika}, 42(02):329--339, 1995.

\bibitem{kriegel2003effective}
H-P Kriegel, Peer Kroger, Zahi Mashael, Martin Pfeifle, Marc{\"o} Potke, and
  Thomas Seidl.
\newblock {Effective similarity search on voxelized CAD objects}.
\newblock In {\em Database Systems for Advanced Applications, 2003.(DASFAA
  2003). Proceedings. Eighth International Conference on}, pages 27--36. IEEE,
  2003.

\bibitem{kriegel2003using}
Hans-Peter Kriegel, Stefan Brecheisen, Peer Kr{\"o}ger, Martin Pfeifle, and
  Matthias Schubert.
\newblock {Using sets of feature vectors for similarity search on voxelized CAD
  objects}.
\newblock In {\em Proceedings of the 2003 ACM SIGMOD international conference
  on Management of data}, pages 587--598. ACM, 2003.

\bibitem{mecke2000additivity}
Klaus~R Mecke.
\newblock {Additivity, convexity, and beyond: applications of Minkowski
  Functionals in statistical physics}.
\newblock In {\em Statistical Physics and Spatial Statistics}, pages 111--184.
  Springer, 2000.

\bibitem{DBLP:journals/corr/NotchenkoKB16}
Alexandr Notchenko, Ermek Kapushev, and Evgeny Burnaev.
\newblock Sparse 3d convolutional neural networks for large-scale shape
  retrieval.
\newblock {\em CoRR}, abs/1611.09159, 2016.

\bibitem{osada2002shape}
Robert Osada, Thomas Funkhouser, Bernard Chazelle, and David Dobkin.
\newblock Shape distributions.
\newblock {\em ACM Transactions on Graphics (TOG)}, 21(4):807--832, 2002.

\bibitem{santalo2004integral}
L.A. Santal{\'o}.
\newblock {\em Integral Geometry and Geometric Probability}.
\newblock Cambridge Mathematical Library. Beijing World Publishing Corporation
  (BJWPC), 2004.

\bibitem{tangelder2008survey}
Johan~WH Tangelder and Remco~C Veltkamp.
\newblock {A survey of content based 3D shape retrieval methods}.
\newblock {\em Multimedia tools and applications}, 39(3):441--471, 2008.

\bibitem{vogel2010quantification}
H-J Vogel, Ulrich Weller, and Steffen Schl{\"u}ter.
\newblock {Quantification of soil structure based on Minkowski functions}.
\newblock {\em Computers \& Geosciences}, 36(10):1236--1245, 2010.

\bibitem{vranic20013d}
Dejan Vranic and Dietmar Saupe.
\newblock {3D shape descriptor based on 3D Fourier transform}.
\newblock In {\em EURASIP}, pages 271--274, 2001.

\bibitem{wu20153d}
Zhirong Wu, Shuran Song, Aditya Khosla, Fisher Yu, Linguang Zhang, Xiaoou Tang,
  and Jianxiong Xiao.
\newblock 3d shapenets: A deep representation for volumetric shapes.
\newblock In {\em Proceedings of the IEEE Conference on Computer Vision and
  Pattern Recognition}, pages 1912--1920, 2015.

\bibitem{zhang2001efficient}
Cha Zhang and Tsuhan Chen.
\newblock {Efficient feature extraction for 2D/3D objects in mesh
  representation}.
\newblock In {\em Image Processing, 2001. Proceedings. 2001 International
  Conference on}, volume~3, pages 935--938. IEEE, 2001.

\bibitem{zhang2007survey}
Lisha Zhang, Manuel~Jo{\~a}o da~Fonseca, Alfredo Ferreira, and Combinando
  Realidade~Aumentada e~Recupera{\c{c}}{\~a}o.
\newblock {Survey on 3D shape descriptors}.
\newblock Technical report.

\end{thebibliography}

\end{document}